\documentclass[final]{l4dc2026}

\title[Test-Time Constraint-Aware Diffusion Predictive Control]{Test-Time Constraint-Aware Diffusion Predictive Control}
\usepackage{times}
\usepackage{bbm}
\usepackage{graphicx}
\usepackage{algorithm}
\usepackage{titletoc}
\usepackage{wrapfig}
\usepackage{float}
\usepackage{url}
\usepackage{lineno,hyperref}
\usepackage{algpseudocode}
\usepackage{amsmath,amssymb,mathtools}

\usepackage{booktabs,pifont}
\theorembodyfont{\itshape}
\theoremheaderfont{\bfseries}
\theorempostheader{}
\theoremsep{}

\theorembodyfont{\upshape}
\theoremheaderfont{\bfseries}

\author{%
 \Name{Braghadeesh Lakshminarayanan}\footnote{Also part of Cyber-physical Systems Group, Aalto University, Espoo, Finland.} \Email{braghadeesh.lakshminarayanan@aalto.fi}\\
 \addr ELLIS Institute Finland
 \AND
 \Name{Ojas Phadake} \Email{ojasphadake@gmail.com }\\
 \addr Indian Institute of Technology, Madras, India%
  \AND
 \Name{Qi Chen} 
 \Email{qi.chen.inf@aalto.fi}\\
 \addr ELLIS Institute Finland \& Aalto University, Espoo, Finland%
 \AND
 \Name{Dominik Baumann} \Email{dominik.baumann@aalto.fi}\\
 \addr Cyber-physical Systems Group, Aalto University, Espoo, Finland
}

\begin{document}

\maketitle

\begin{abstract}%
Diffusion models can learn multimodal distributions over
finite-horizon state-input sequences, providing a way to
generate candidates for predictive control. However,
sequences generated by a pretrained model may violate
safety constraints introduced at deployment, for example
when obstacles are added or repositioned. Retraining
whenever these constraints change limits the usefulness
of learned predictive controllers. In this paper, we
propose \emph{Test-Time Constraint-Aware Diffusion Predictive
Control} (TCA-DPC), an algorithm for adapting the sampling
distribution of a pretrained diffusion controller to newly
introduced safety constraints without retraining.
We treat the pretrained model as a prior over finite-horizon
sequences and define an adapted distribution through
exponential tilting based on a constraint-violation penalty.
The tilt strength is adjusted online using the fraction
of generated sequences that fail constraint verification.
We approximately sample the adapted distribution using
the Metropolis-adjusted Langevin algorithm in the
Gaussian noise space of the pretrained diffusion sampler.
 
\end{abstract}


\section{Introduction}

In predictive control, multiple input sequences may achieve the same objective while exhibiting different behaviours. For example, different initializations of a nonlinear model predictive control solver may yield distinct locally optimal solutions~\citep{huang2024toward,julbe2025diffusion}, while human demonstrations may follow different routes around an obstacle. Such variations can give rise to multimodal distributions over control sequences or joint state-input trajectories. Diffusion models provide a way to learn these distributions from data obtained through numerical optimization, simulation, or human demonstrations~\citep{pmlr-v162-janner22a,urain2024deep}. Once trained, the model generates finite-horizon sequences through a reverse denoising process, allowing different modes of the learned distribution to be sampled. In what follows, we refer to the trained diffusion model together with its reverse denoising procedure as the \emph{pretrained diffusion sampler}.

The distribution induced by the pretrained sampler reflects the behaviours and constraints represented in the training data. At deployment, additional safety requirements may arise, for example when a new obstacle occupies a previously traversable region.
Sequences that remain likely under the pretrained sampler may then violate the deployment constraints. Retraining the diffusion model whenever these requirements change is costly. We therefore seek to
adapt the pretrained sampler to the deployment constraints while keeping the trained model parameters fixed.

Existing approaches modify diffusion sampling through
guidance~\citep{bansal2024universal,guo2024gradient} or
projection~\citep{christopher2024constrained}. In robot motion planning, Motion Planning Diffusion uses gradient guidance to incorporate planning objectives and previously unseen obstacles~\citep{carvalho2023motion}.
For predictive
control, \citet{romer2024diffusion} propose Diffusion
Predictive Control with Constraints (DPCC), which incorporates
model-based projections into the DDPM denoising
process~\citep{ho2020denoising}. Their experiments demonstrate
improved constraint satisfaction and task performance over
the guidance and projection baselines considered.
DPCC formulates these projections using an explicit
description of the state and input constraints, including
obstacle geometry in its navigation experiments. It also
tightens the state constraints using a model-mismatch bound, effectively adding clearance around obstacles.
Its robustness guarantee depends on the validity of this
bound. These requirements motivate an alternative that accesses
deployment constraints through an evaluator providing
violation penalties and their derivatives, without requiring
an explicit geometric description of the feasible set,
a projection operator, or a prescribed model-mismatch bound.

In this paper, we develop \emph{Test-Time Constraint-Aware Diffusion Predictive Control} (TCA-DPC), a gradient-based procedure for adapting a pretrained diffusion sampler to new safety constraints in
predictive control. We take the distribution induced by the pretrained sampler as the \emph{pretrained diffusion prior} and reweight it through a change of measure via an exponential weight, also known as \emph{exponential tilting}, which
admits an interpretation as loss-based updating of the prior distribution~\citep{bissiri2016general}. The exponential weight depends
on a penalty for violating the deployment constraints, so sequences
with larger penalties receive smaller weights. All sequences with
zero penalty receive the same weight before normalization.
Consequently, the tilted target preserves the relative probabilities of feasible behaviours whenever the prior assigns positive probability to feasible sequences. We implement this adaptation by changing the distribution of the random inputs supplied to the pretrained sampler. The trained denoising network remains unchanged. This allows us to use derivatives of the violation penalty through the complete denoising
process to construct proposals in the noise domain. We also adjust the tilt strength during control using feedback from the verification of generated sequences. Our contributions are as follows:
\begin{itemize}
\item We construct a target distribution in noise space whose image under the pretrained diffusion sampler is the desired exponentially tilted distribution over sequences. We approximately sample this target using the Metropolis-adjusted Langevin algorithm (MALA)~\citep{roberts2002langevin}, with proposals based on gradients through the complete sampler.

\item We introduce an online feedback rule for the tilt strength. At each control step, the rule compares the observed fraction of generated sequences that fail verification with a prescribed target failure rate. It increases or decreases the tilt strength accordingly, within a specified range, without updating the pretrained diffusion model.
\end{itemize}


\paragraph{Related work.}
Exponential tilting has also been used to adapt diffusion-generated distributions in other settings. DiffTilt~\citep{khandait2026diffusion} reweights a learned joint distribution over environments and system
executions to search for safety-critical failures. Our objective is instead to generate sequences with fewer violations of deployment constraints. BayesFP~\citep{sirigiri2026bayesfp} formulates
inference-time adaptation of pretrained diffusion and flow models as posterior sampling with a cost-derived likelihood. It builds on the Feynman-Kac corrector
framework~\citep{skreta2025feynman} to approximate the resulting distribution. Concurrent work, MOSAIC~\citep{bodmer2026mosaic}, also adapts a distribution over the stochastic inputs of a pretrained diffusion model and adjusts constraint penalties using feedback. It uses gradient-free, importance-weighted updates of mode-specific proposal
distributions. Our approach uses derivatives of the violation penalty through the complete diffusion sampler to construct MALA proposals in noise space. The feedback also differs: MOSAIC updates its constraint multipliers using importance-weighted estimates of constraint costs, whereas our update uses the observed fraction of sequences that fail verification across successive control steps.

\section{Problem Setup}
\label{sec:prelims}

Consider the discrete-time dynamical system
\begin{equation}\label{eq:model}
s_{t+1}=f(s_t,u_t),
\qquad
f:\mathbb R^n\times\mathbb R^m\to\mathbb R^n,
\end{equation}
where $s_t$ is the state and $u_t$ is the applied
control input. The state-transition function $f$
is time-invariant and need not be known exactly. The control objective is to steer the system toward
a prescribed target while satisfying state and input constraints, including safety requirements introduced
at deployment. In conventional model predictive control~\citep{Borrelli_Bemporad_Morari_2017}, this objective
is addressed by planning, at each control time $t$, future inputs and states over a prediction horizon of $N$ stages through a finite-horizon optimization
using an approximate dynamics model $\hat f$.


Here, a pretrained diffusion model generates candidate
state-input sequences for this planning step. In particular, a diffusion model is trained
on finite-horizon segments of offline expert demonstrations (similar to \emph{imitation learning}~\citep{osa2018algorithmic}),
it approximates their conditional distribution, which
can represent different ways of reaching the target.
At control time $t$, a generated sequence takes the form
\begin{equation}\label{eq:sequence}
Y=
\big(
(\hat s_{0\mid t},\hat u_{0\mid t}),
\ldots,
(\hat s_{N-1\mid t},\hat u_{N-1\mid t})
\big)
\in\mathcal Y\subseteq\mathbb R^d,
\end{equation}
where $d=N(n+m)$. The quantities $\hat s_{k\mid t}$
and $\hat u_{k\mid t}$ denote the generated state
prediction and planned input for time $t+k$ respectively, based
on the information available at time $t$, with
$\hat s_{0\mid t}=s_t$.

Let $c_t$ denote the conditioning information, including
the current state and any task information used by the
sampler. For example, in a navigation task, this may specify a goal line, with the episode ending when
the system reaches it. For fixed $c_t$, the pretrained sampler,
driven by fresh Gaussian random inputs, generates
\begin{equation}\label{eq:diffusion_prior}
Y\sim\mathbb P_t,
\qquad
\mathbb P_t(\cdot)
:=
\mathbb P_\theta(\cdot\mid c_t),
\end{equation}
where $\theta$ denotes the trained parameters.
We call $\mathbb P_t$ the \emph{pretrained diffusion prior}.

The controller operates in a receding-horizon manner:
at each control step, it selects a sequence from
the generated candidates and applies only its first
input, $u_t=\hat u_{0\mid t}$. After observing $s_{t+1}$,
it generates new sequences and repeats the procedure
until the episode's termination condition is met. The episode length $T$ counts the executed control
steps and may vary between episodes; it need not
equal the prediction horizon $N$. An individual
generated sequence therefore need not reach the
target within its prediction horizon. Our objective is to adapt the sampling distribution
to the deployment constraints while keeping the
trained parameters $\theta$ of the diffusion model fixed.

\paragraph{Diffusion model.}
We use a pretrained denoising diffusion probabilistic model
(DDPM)~\citep{ho2020denoising}. During training, sequences are
corrupted with Gaussian noise, and a neural network learns to
predict the added noise. Generation proceeds through a learned
reverse denoising process. Let $Y^{(i)}$ denote the sequence at diffusion step $i$, with
$K$ denoising steps and $Y^{(0)}$ the generated sequence.
Sampling starts from $Y^{(K)}\sim\mathcal N(0,I_d)$ before
conditioning. Each reverse update, before reimposing the
conditioning values, takes the form
\(
Y^{(i-1)}
\leftarrow
\mu_{\theta,i}(Y^{(i)})+\sigma_i\xi_i,
\,\,i=K,\ldots,1,
\) where $\mu_{\theta,i}$ is the DDPM reverse mean computed from
the noise-prediction network and the
prescribed noise schedule~\citep{pmlr-v162-janner22a}. The variables
$\xi_i\sim\mathcal N(0,I_d)$ are independent of one another
and of $Y^{(K)}$. The sampling noise scales $\sigma_i$ are
prescribed, with $\sigma_1=0$. In this paper, conditioning on the current state is
enforced by fixing the corresponding state entries at
initialization and after every reverse update. These operations
are determined by $c_t$. The resulting sequence $Y=Y^{(0)}$
has distribution $\mathbb P_\theta(\cdot\mid c_t)$.

\paragraph{Test-time safety constraints.}
The pretrained diffusion prior reflects the behaviours and constraints represented in
the training data. At deployment, additional requirements may
arise, for example when an obstacle occupies a previously
traversable region. These requirements can exclude sequences
that remain likely under the pretrained model. We express the requirements through signed
constraint residuals
\begin{equation}\label{eq:new_constraints}
g_{\ell,t}(Y)\leq0,
\qquad \ell=1,\ldots,e,
\end{equation}
where $e \in \mathbb{N}$, and \(g_{\ell,t}\colon \mathcal{Y} \to \mathbb{R}\). 
The index $\ell$ enumerates the constraints over all relevant
prediction stages. Their evaluation depends on the quantities
contained in $Y$ and the prediction procedure used by the
underlying controller. Dependence on the current state and
other information in $c_t$ is implicit in the index $t$.
The residuals may also include nominal admissibility and
model-consistency requirements. Any constraint tightening is
incorporated into their definition. The corresponding feasible set is \(
\mathcal A_t
=
\left\{
Y\in\mathcal Y:
g_{\ell,t}(Y)\leq0
\text{ for every }\ell=1,\ldots,e
\right\}.
\) 
We assume access to a constraint evaluator that provides
the violation penalty and the derivatives required by
the sampling procedure. The adaptation does not require
a projection operator onto $\mathcal A_t$ or an explicit
description of its boundary. This distinction
between oracle access and explicit constraint representations
is standard in constrained convex optimization; see, for example,
\cite{pmlr-v75-lee18a,pmlr-v178-mhammedi22a}. To quantify violations, let
$\rho:\mathbb R\to\mathbb R_{\geq0}$ be a continuously
differentiable, non-decreasing function satisfying
\(
\rho(a)=0\quad\text{for }a\leq0,
\,
\rho(a)>0\,\text{for }a>0.
\)
Define
\begin{equation}\label{eq:safety_penalty}
\Phi_t(Y)
=
\sum_{\ell=1}^{e}
\rho\!\left(g_{\ell,t}(Y)\right).
\end{equation}
A suitable choice for $\rho$ is the squared hinge
$\rho(a)=[a]_+^2$, where $[a]_+=\max\{a,0\}$. This construction gives
\begin{equation}\label{eq:penalty_exactness}
\Phi_t(Y)\geq0,
\qquad
\Phi_t(Y)=0
\quad\Longleftrightarrow\quad
Y\in\mathcal A_t.
\end{equation}
We assume that $\Phi_t$ is measurable and finite
$\mathbb P_t$-almost surely; an upper bound on the penalty
is not required.

\paragraph{Adaptation objective.}
Given the pretrained diffusion prior $\mathbb P_t$ and the deployment
constraints, our objective is to construct and approximately
sample an adapted distribution $\mathbb Q_t$ that increases
the probability of constraint-satisfying sequences. The adaptation must operate within a prescribed sampling
budget at each control step and use feedback from sequence
verification without changing the trained parameters $\theta$.
The following section constructs an exponentially tilted
family $\mathbb Q_{\beta,t}$ and develops an online rule
for adjusting its tilt strength $\beta$.



\section{Test-Time Constraint-Aware Diffusion Predictive Control}\label{sec:proposed}

We adapt the pretrained diffusion prior by assigning smaller weights to sequences that violate the deployment constraints. The adaptation retains the relative prior probabilities of feasible sequences and leaves the trained parameters unchanged. We first define the adapted distribution, known as the \emph{diffusion posterior}, then describe how to sample it approximately through the random inputs of the pretrained diffusion sampler. The adaptation strength, called the \emph{tilt parameter}, is selected online using feedback from constraint verification. The proposed approach (TCA-DPC) combines three components: exponential tilting of the pretrained diffusion prior, approximate sampling through noise-space MALA, and an online feedback rule for adjusting the tilt strength. 


\subsection{Exponential tilting of the pretrained diffusion prior: The diffusion posterior}

The diffusion posterior should depend on the safety violations associated with the generated sequences. We capture this dependence through a multiplicative factor that reweights the pretrained diffusion prior according to the violation penalty. To formalize this adaptation, we use a standard change-of-measure approach from probability theory, namely exponential tilting~\citep{bissiri2016general}. For a tilting parameter \(\beta > 0\), the diffusion posterior is defined as 

\begin{equation}\label{eq:posterior_tilt}
 \mathbb Q_{\beta,t}(dY)
 =
 \frac{\exp[-\beta\Phi_t(Y)]}{\mathcal Z_{\beta,t}}\,
 \mathbb P_t(dY),
 \qquad
 \mathcal Z_{\beta,t}
 =
 \mathbb E_{\mathbb P_t}
 \big[\exp[-\beta\Phi_t(Y)]\big].
\end{equation}
Since $\Phi_t$ is nonnegative and finite, $0<\mathcal Z_{\beta,t}\leq1$ for every finite $\beta$.
Increasing $\beta$ strengthens the preference for sequences with smaller violations, while
$\mathbb Q_{0,t}=\mathbb P_t$.
The purpose of $\mathbb Q_{\beta,t}$ is to reduce the probability of constraint
violations while retaining the different feasible behaviours represented by the prior. The following proposition captures this notion. 

\begin{proposition}[Effect of exponential tilting]
\label{prop:tilt}
For a fixed control time $t$ with $\mathbb P_t(\mathcal A_t)>0$,
the tilted distributions defined in~\eqref{eq:posterior_tilt}
satisfy
$\mathbb Q_{\beta,t}(\,\cdot\mid\mathcal A_t)
=\mathbb P_t(\,\cdot\mid\mathcal A_t)$
for every finite $\beta\geq0$, while the violation probability
$\mathbb Q_{\beta,t}(\mathcal A_t^c)$ is non-increasing in $\beta$
and converges to zero as $\beta\to\infty$.
\end{proposition}

\noindent The proof of Proposition~\ref{prop:tilt} follows from straightforward arguments, but for completeness, we have added it to the supplementary material. Proposition~\ref{prop:tilt} gives a precise interpretation of the
adaptation: increasing $\beta$ reduces the violation probability
without changing the relative probabilities of feasible sequences. 
These properties motivate adjusting $\beta$ online using feedback from constraint verification. Before updating $\beta$ online, we now describe how to draw samples from the diffusion posterior.

\subsection{Sampling from the diffusion posterior}
\label{subsec:posterior_sampling}

To sample from the diffusion posterior, we express the pretrained diffusion sampler as a deterministic
function of its Gaussian random inputs. Collect the initial draw
$Y^{(K)}$, before conditioning, and the independent reverse-step
noises into
\(
z=(Y^{(K)},\xi_K,\ldots,\xi_2)\in\mathbb R^{d_z},
\qquad
z\sim\Gamma:=\mathcal N(0,I_{d_z}),
\)
omitting noises at steps with $\sigma_i=0$. For fixed $c_t$,
the complete denoising procedure, including conditioning and
noise scaling, defines a map $\Psi_{c_t}$ such that
$Y=\Psi_{c_t}(z)$. Thus, the pretrained diffusion prior $\mathbb P_t$
is the pushforward of $\Gamma$ under $\Psi_{c_t}$. For every measurable $A\subseteq\mathcal Y$, the pushforward
integration formula~\citep{bogachev2007measure} gives
\begin{equation}
\label{eq:q_beta_in_terms_of_transformed_gaussian}
\begin{aligned}
\mathbb Q_{\beta,t}(A)
&=
\frac{1}{\mathcal Z_{\beta,t}}
\int_{\mathcal Y}
\mathbf1_A(Y)e^{-\beta\Phi_t(Y)}\,\mathbb P_t(dY)
=
\frac{1}{\mathcal Z_{\beta,t}}
\int_{\Psi_{c_t}^{-1}(A)}
e^{-\beta\Phi_t(\Psi_{c_t}(z))}\,\Gamma(dz),
\end{aligned}
\end{equation}
where $\Psi_{c_t}^{-1}(A)=\{z:\Psi_{c_t}(z)\in A\}$.
Defining \(
\pi_{\beta,t}(dz)
=
\frac{
e^{-\beta\Phi_t(\Psi_{c_t}(z))}
}{
\mathcal Z_{\beta,t}
}\,\Gamma(dz)\),
we obtain
$\mathbb Q_{\beta,t}(A)
=\pi_{\beta,t}(\Psi_{c_t}^{-1}(A))$.
Consequently, sampling $z\sim\pi_{\beta,t}$ and applying
$\Psi_{c_t}$ produces a sequence with law $\mathbb Q_{\beta,t}$,
without evaluating the density of $\mathbb P_t$ or requiring
$\Psi_{c_t}$ to be invertible. The density of $\pi_{\beta,t}$ is proportional to
$\exp[-V_{\beta,t}(z)]$, with
\begin{equation}\label{eq:noise_potential}
V_{\beta,t}(z)
=
\frac12\|z\|^2
+
\beta\Phi_t(\Psi_{c_t}(z)).
\end{equation}
We use the Metropolis-adjusted Langevin algorithm
(MALA)~\citep{roberts2002langevin,shi2025metropolis} to
approximately sample this target. Gradients are computed
through the complete diffusion sampler and the violation penalty. Holding $c_t$ and $\beta$ fixed, iteration
$j=0,\ldots,M-1$ proposes
\begin{equation}\label{eq:mala_proposal}
z'=z^{(j)}- h\nabla_z V_{\beta,t}(z^{(j)}) + \sqrt{2h}\,\zeta_j,
\qquad
\zeta_j\sim\mathcal N(0,I_{d_z}),
\end{equation}
where $h>0$ is fixed and the $\zeta_j$ are independent.
Let $q_{\beta,t}(z'\mid z)$ denote the Gaussian proposal density
with mean $z-h\nabla_z V_{\beta,t}(z)$ and covariance $2hI_{d_z}$.
The proposal is accepted with probability
\begin{equation}\label{eq:mala_acceptance}
a_{\beta,t}(z,z')
=
\min\left\{
1,\,
e^{V_{\beta,t}(z)-V_{\beta,t}(z')}
\frac{q_{\beta,t}(z\mid z')}{q_{\beta,t}(z'\mid z)}
\right\}.
\end{equation}
If accepted, $z^{(j+1)}=z'$; otherwise,
$z^{(j+1)}=z^{(j)}$. The normalizing constant cancels from
the acceptance ratio. For a fixed target, this Metropolis correction leaves
$\pi_{\beta,t}$ invariant. After $M$ iterations, we generate
$Y=\Psi_{c_t}(z^{(M)})$ as an approximate sample from
$\mathbb Q_{\beta,t}$.

This construction provides an approximate sampler for $\mathbb Q_{\beta,t}$. It remains to select the tilt strength
during deployment. Small values of $\beta$ leave the target distribution close to the pretrained diffusion prior, with
$\mathbb Q_{0,t}=\mathbb P_t$, and may provide insufficient preference for constraint-satisfying sequences. Larger values strengthen this preference but can make the target harder to explore. We therefore adjust the tilt strength online using feedback from sequence verification.
At each control step $t$, we hold $\beta_t$ fixed while generating a batch of $B$ sequences and use the fraction that
fail verification to determine $\beta_{t+1}$. The following subsection presents this feedback rule.

\subsection{Online selection of the tilt parameter}
\label{subsec:tilt_parameter_selection}

We adjust the tilt strength using verification feedback,
aiming to regulate the fraction of failing sequences
around a prescribed \emph{maximum failure rate}
$\delta\in[0,1]$. At control time $t$, we hold $\beta_t$
fixed and generate $B$ sequences
$\{Y_t^{(b)}\}_{b=1}^{B}$ using MALA targeting
$\mathbb Q_{\beta_t,t}$. After the initial $M$ MALA
steps, we compute the failure fraction and update
the tilt for the next control step:
\begin{align}
f_t
&=
\frac{1}{B}\sum_{b=1}^{B}
\mathbf{1}\!\left\{Y_t^{(b)}\notin\mathcal A_t\right\},
\label{eqref:fraction_failure}\\
\beta_{t+1}
&=
\operatorname{clip}_{[0,\beta_{\max}]}
\bigl(\beta_t+\kappa(f_t-\delta)\bigr),
\label{eq:beta_update}
\end{align}
where $\kappa>0$ is the feedback gain,
$\beta_{\max}>0$ bounds the tilt strength, and
$\operatorname{clip}$ projects its argument onto the
indicated interval. We initialize
$\beta_0\in[0,\beta_{\max}]$ and measure $f_t$ before
any additional refinement or fallback. The update implements projected integral feedback:
it accumulates the discrepancy $f_t-\delta$, increasing
the tilt above the maximum failure rate and decreasing it
below the budget, subject to the bounds.
This direction is motivated by
Proposition~\ref{prop:tilt}: for a fixed context and
constraint set, the violation probability under the
ideal tilted distribution is non-increasing in $\beta$.
The cap $\beta_{\max}$ is a computational design choice,
since stronger tilting can make the target harder to
explore within a finite MALA budget. 


\subsection{Receding-horizon implementation}
\label{subsec:control_implementation}

The preceding subsections define the adapted sampler
and its feedback update. For control execution, the controller selects a verified sequence minimizing
an application-dependent cost $J_{\mathrm{sel}}$, such
as deviation from the previously selected sequence,
and applies its first input. A finite sampling budget
may produce no verified sequence and therefore, a fallback is necessary. Selecting a failed
sequence at random ignores the severity of
its predicted violations and could direct the system
toward an unsafe region. We therefore first attempt a
bounded number of additional MALA rounds at the current
control step. Each round updates all $B$ noise vectors
once, with $\beta_t$ and $h$ fixed, and checks the
regenerated sequences. Refinement stops when a sequence
passes verification, and selection proceeds using
$J_{\mathrm{sel}}$.

If no sequence passes, the controller continues the
previously selected sequence by applying its next planned
input that has not yet been applied: for example, after
applying $\hat u_{0\mid t-1}$ at time $t-1$, reuse at
time $t$ applies $\hat u_{1\mid t-1}$. If that sequence
was verified, the constraints remain unchanged, and
tracking is accurate, its remaining inputs provide a
previously checked continuation. Reuse is permitted for a
limited number of consecutive control steps, during which
ordinary MALA refinement and tilt updates continue. This
limit bounds reliance on earlier predictions. If no
previous sequence is available or the reuse limit is
reached, the controller applies the first input of the
sequence with the smallest maximum constraint violation.
These fallback choices are heuristic
and do not guarantee eventual verification success.

Algorithm~\ref{alg:Tilt_Diffusion} summarizes the controller.
The stored sequence $Y_{\mathrm{prev}}$ contains the
remaining state-input pairs of the most recently selected
sequence; each pair is removed after its input is applied.
Initially, all noise vectors are independent draws from
$\mathcal N(0,I_{d_z})$, $Y_{\mathrm{prev}}$ is empty, and
$n_{\mathrm{reuse}}=0$. This counter increases after each
reuse and resets whenever a newly generated sequence
is selected.

\begin{algorithm}[h!]
\fontsize{7}{7}\selectfont
\SetAlFnt{\fontsize{8}{9}\selectfont}
\SetAlCapFnt{\footnotesize}
\SetAlCapNameFnt{\footnotesize}
\SetInd{0.4em}{0.6em}
\DontPrintSemicolon

\caption{TCA-DPC}
\label{alg:Tilt_Diffusion}

\KwIn{Context $c_t$; stored noise vectors
$\{z^{(b)}\}_{b=1}^{B}$, tilt $\beta_t$,
remaining previous sequence $Y_{\mathrm{prev}}$,
and reuse counter $n_{\mathrm{reuse}}$;
$J_{\mathrm{sel}},M,h,\delta,\kappa,\beta_{\max}$;
reuse limit $R_{\max}$ and additional refinement
budget $M_{\mathrm{extra}}$}

\tcp{Sampling and feedback}
Construct $\Phi_t$ and $V_{\beta_t,t}$ using
Eq.~\eqref{eq:noise_potential}\;

Refresh one $z^{(b)}\sim\mathcal N(0,I_{d_z})$;
retain the other noise vectors\;

\For{$b=1,\ldots,B$}{
  Perform $M$ MALA updates of $z^{(b)}$ using
  Eqs.~\eqref{eq:mala_proposal}--\eqref{eq:mala_acceptance}
  with $\beta=\beta_t$ and step size $h$\;
  $Y^{(b)}\gets\Psi_{c_t}(z^{(b)})$\;
}

$\mathcal I\gets\{b:Y^{(b)}\text{ passes verification}\}$\;
$f_t\gets1-|\mathcal I|/B$
\tcp*{measured before fallback}

$\beta_{t+1}\gets
\operatorname{clip}_{[0,\beta_{\max}]}
\bigl(\beta_t+\kappa(f_t-\delta)\bigr)$\;

\tcp{Control selection and fallback}
$j\gets0$\;
\While{$\mathcal I=\varnothing$ and $j<M_{\mathrm{extra}}$}{
  Perform one further MALA update for each $z^{(b)}$
  with the same $\beta_t$ and $h$;
  regenerate $Y^{(b)}=\Psi_{c_t}(z^{(b)})$\;, 
  $\mathcal I\gets
  \{b:Y^{(b)}\text{ passes verification}\}$;
  $j\gets j+1$\;
}

\uIf{$\mathcal I\neq\varnothing$}{
  $b^\star\gets
  \arg\min_{b\in\mathcal I}J_{\mathrm{sel}}(Y^{(b)})$;
  $Y_{\mathrm{sel}}\gets Y^{(b^\star)}$;
  $n_{\mathrm{reuse}}\gets0$\;
}
\uElseIf{$Y_{\mathrm{prev}}$ is nonempty and
$n_{\mathrm{reuse}}<R_{\max}$}{
  $Y_{\mathrm{sel}}\gets Y_{\mathrm{prev}}$;
  $n_{\mathrm{reuse}}\gets n_{\mathrm{reuse}}+1$\;
}
\Else{
  $b^\star\gets
  \arg\min_{b=1,\ldots,B}
  \max_{\ell=1,\ldots,e}g_{\ell,t}(Y^{(b)})$;
  $Y_{\mathrm{sel}}\gets Y^{(b^\star)}$;
  $n_{\mathrm{reuse}}\gets0$
  \tcp*{unverified selection}
}

Apply the first input of $Y_{\mathrm{sel}}$ and set
$Y_{\mathrm{prev}}$ to its remaining sequence\;

Store $\{z^{(b)}\}_{b=1}^{B}$, $\beta_{t+1}$,
$Y_{\mathrm{prev}}$, and $n_{\mathrm{reuse}}$\;
\end{algorithm}

\section{Numerical Study}
\label{sec:numerical}

We evaluate\footnote{Our code is available at \url{https://github.com/ojasph/tca-dpc}.} TCA-DPC (Algorithm~\ref{alg:Tilt_Diffusion}) on the
\emph{Avoiding} task from D3IL~\citep{jia2024towards},
using the test-time constraint setting of
DPCC~\citep{romer2024diffusion}. Using the same pretrained
models, we compare adaptive tilting with projection-based,
untilted, and fixed-tilt baselines. We assess constraint
violations and closed-loop performance, track the tilt
strength throughout each episode,
and examine sensitivity to constraint tightening.

\begin{figure}[h]
  \centering
  \includegraphics[width=0.8\linewidth]{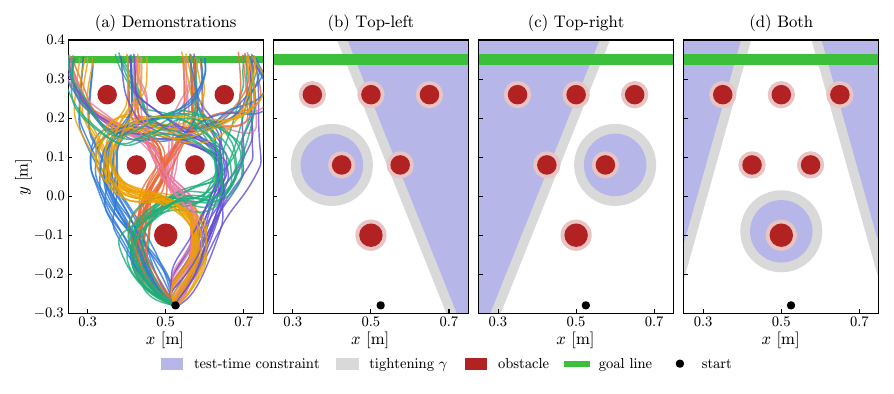}
  \vspace{-2 em}
  \caption{Avoiding environment. (a) The $96$ training demonstrations, coloured by route, i.e., by the side on which they pass the first obstacle and the gap they take between the two middle obstacles. (b)--(d) The three test-time constraint layouts: forbidden regions (blue) and their tightening by $\gamma$ (grey). None of the forbidden regions appears in the training data.}
  \label{fig:env}
\end{figure}

\paragraph{Environment, data, and test-time constraints.}
In \emph{Avoiding}, a simulated manipulator moves its
end-effector to a goal line while avoiding six obstacles.
Following DPCC~\citep[Section~6.1]{romer2024diffusion},
$s_t\in\mathbb R^4$ contains the current and desired
planar end-effector positions, and $u_t\in\mathbb R^2$
is the desired Cartesian velocity supplied to a low-level
controller. The diffusion model is trained on $96$ human
demonstrations: four for each of $24$ routes, to generate
joint state-input sequences and remains frozen during evaluation. It generates $Y$ with $N=8$ stages, conditioned on $c_t=s_t$.
Only the first planned velocity $\hat u_{0\mid t}$ is sent to the
low-level controller, with replanning at every step.
Since the demonstrations cover $24$ routes, $\mathbb P_t$ is
multimodal, and tilting suppresses routes blocked by the new
obstacles. Deployment introduces additional forbidden regions absent
during training. We use three layouts: \emph{top-left}
(one halfspace and one circle), \emph{top-right}
(its reflection about $x=0.5$), and \emph{both}
(two halfspaces and one circle). We compare TCA-DPC and DPCC without constraint tightening
($\gamma=0$) and with a buffer $\gamma>0$, which shifts
halfspace boundaries and enlarges circular obstacles.
Figure~\ref{fig:env}
illustrates the demonstrations and constraint layouts.
The constraint residuals $g_{\ell,t}(Y)\leq0$ specify obstacle
avoidance, input bounds derived from the demonstrations,
and consistency with an approximate dynamics model.
Their explicit expressions, the sequence representation,
and the implementation of the penalty $\Phi_t$ are
provided in the supplementary material. We implement Algorithm~\ref{alg:Tilt_Diffusion} with
pretrained diffusion models, selecting the verified sequence closest to the unexecuted portion of the previous plan. Further, we set the consecutive reuse limit to $R_{\max}=3$
and the additional refinement budget to
$M_{\mathrm{extra}}=10$. These limits were selected
by manual tuning in preliminary simulations,
considering closed-loop performance and computational cost. Model and training details, sampling settings, verification tolerances, and feedback parameters are provided in the supplementary material. 

\paragraph{Baselines and metrics.}
All methods use the same pretrained diffusion models.
\emph{Prior} samples without verification, as in
Diffuser~\citep{pmlr-v162-janner22a}.
\emph{Prior + verification} is Alg.~\ref{alg:Tilt_Diffusion} with
$M=M_{\mathrm{extra}}=0$ and the same warm start (one fresh $z^{(b)}$
per step), verification, selection, and $Y_{\mathrm{prev}}$ reuse;
its candidates are thus step-wise correlated samples of $\mathbb P_t$,
and it differs from TCA-DPC only in omitting MALA refinement.
\emph{Fixed $\beta$} uses $\beta=\beta_0$ ($\kappa=0$). DPCC-C and DPCC-T select
sequences using cumulative projection cost and temporal
consistency, respectively (see~\cite{romer2024diffusion} for details).
The maximum episode length is $200$ control steps.
Within this limit, we report goal success (G), defined as the fraction of episodes in which the end-effector reaches the goal line, and goal-and-constraint success (G\&C), the fraction that reach the goal line without violating the original state constraints at any control step. Constraint satisfaction is evaluated without the
tightening buffer, irrespective of the value of $\gamma$ used by the controller. We also report the mean number of violating control steps per episode (Viol.) and the mean number of steps to the goal over goal-reaching episodes (Steps).
For TCA-DPC and its baselines, FB is the fraction of control steps that reach the fallback branch of Alg.~\ref{alg:Tilt_Diffusion} (no verified sequence after the $M_{\mathrm{extra}}$ additional updates), and $\bar f$ is the mean of $f_t$; both are pooled over each model's control steps and averaged over models.
Each layout is evaluated using five trained models and ten episodes per model, with standard errors computed across models. Each run uses an NVIDIA V100, A100 or H100 GPU.

\vspace{-1 em}

\subsection{Results}\label{subsec:results}
\begin{figure}[t]
    \centering
    \includegraphics[width=0.8\linewidth]{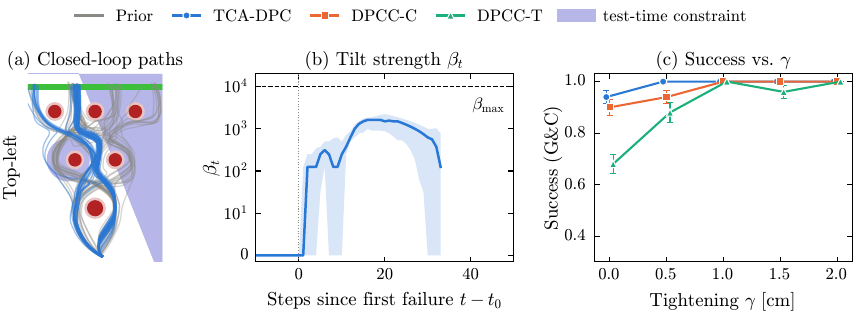}
    \caption{Top-left layout. (a) Closed-loop paths of the prior (grey) and TCA-DPC (blue) without tightening. (b) Tilt strength $\beta_t$ after the first verification failure (median and IQR). (c) Goal-and-constraint success versus tightening $\gamma$.}
    \label{fig:paper_scenario_tl}
\end{figure}

We evaluate the adapted sampler in closed loop on the three constraint layouts, comparing success and constraint violations against the untilted (pretrained diffusion prior), fixed-tilt (constant $\beta$), and DPCC baselines. We then examine how the feedback rule adjusts the tilt strength during an episode, and how performance changes with constraint tightening $\gamma$.
Table~\ref{tab:main} compares the controllers without
constraint tightening. Averaged over the three layouts,
TCA-DPC achieves $82.7\%$ G\&C,
compared with $61.3\%$ for the untilted verification
baseline and $71.3\%$ for DPCC-C. The mean number of
violating steps per episode decreases from $3.17$ for
the untilted verification baseline to $0.73$ for our
method. The largest difference from DPCC-C occurs in
the \emph{both} layout, where TCA-DPC achieves
$74\%$ success compared with $46\%$. These results are obtained without projection or constraint tightening ($\gamma=0$). Across the three layouts, TCA-DPC achieves the highest
or joint-highest mean G\&C among the evaluated controllers.
Figure~\ref{fig:paper_scenario_tl}(a) compares the
closed-loop end-effector paths obtained with the pretrained
sampler and TCA-DPC in the \emph{top-left} layout,
where G\&C increases from $10\%$ to $94\%$.
\begin{table}[h]
\centering
\caption{ Closed-loop performance on the three constraint layouts without tightening ($\gamma=0$); Viol., Steps and FB are averaged over layouts. Bold marks the best result and underlining the best among methods without projection.}
\label{tab:main}
\footnotesize
\setlength{\tabcolsep}{2.4pt}
\renewcommand{\arraystretch}{0.95}

\begin{tabular}{@{}lc cc cc cc ccc@{}}
\toprule
 & & \multicolumn{2}{c}{Top-left}
   & \multicolumn{2}{c}{Top-right}
   & \multicolumn{2}{c}{Both} & & & \\
\cmidrule(lr){3-4}
\cmidrule(lr){5-6}
\cmidrule(lr){7-8}
Method & Proj.
& G & G\&C & G & G\&C & G & G\&C
& Viol.\,$\downarrow$ & Steps\,$\downarrow$ & FB \\
\midrule
Prior
& \ding{55}
& 1.00 & 0.10{\scriptsize$\pm$.03}
& 1.00 & 0.06{\scriptsize$\pm$.02}
& 1.00 & 0.02{\scriptsize$\pm$.02}
& 16.66 & 74.8 & -- \\

Prior + verification
& \ding{55}
& 1.00 & 0.90{\scriptsize$\pm$.00}
& 0.80 & 0.46{\scriptsize$\pm$.06}
& 0.94 & 0.48{\scriptsize$\pm$.07}
& 3.17 & 73.1 & 0.12 \\

Fixed $\beta=\beta_0$
& \ding{55}
& 1.00 & \textbf{0.94}{\scriptsize$\pm$.04}
& 0.60 & 0.38{\scriptsize$\pm$.09}
& 0.96 & 0.62{\scriptsize$\pm$.11}
& 2.12 & 69.1 & 0.08 \\

\textbf{TCA-DPC}
& \ding{55}
& 1.00 & \textbf{0.94}{\scriptsize$\pm$.04}
& 0.86 & \textbf{0.80}{\scriptsize$\pm$.05}
& 0.94 & \textbf{0.74}{\scriptsize$\pm$.07}
& \textbf{0.73} & \underline{68.8} & 0.05 \\
\midrule
DPCC-C
& \ding{51}
& 0.98 & 0.90{\scriptsize$\pm$.03}
& 0.94 & 0.78{\scriptsize$\pm$.07}
& 0.90 & 0.46{\scriptsize$\pm$.11}
& 0.89 & 66.9 & -- \\

DPCC-T
& \ding{51}
& 0.94 & 0.68{\scriptsize$\pm$.04}
& 0.94 & 0.68{\scriptsize$\pm$.07}
& 0.86 & 0.56{\scriptsize$\pm$.11}
& 1.85 & \textbf{65.7} & -- \\
\bottomrule
\end{tabular}
\end{table}
Figure~\ref{fig:paper_scenario_tl}(b) illustrates online
adaptation in the \emph{top-left} layout without tightening.
Following the first verification failure, the median
tilt strength rises to the order of $10^3$ and subsequently
decreases. These values emerge from verification feedback
during execution. Selecting a suitable fixed strength
beforehand may instead require separate calibration. Holding the tilt fixed at $\beta=\beta_0=10$ instead lowers the average goal-and-constraint success to $64.7\%$ (Table~\ref{tab:main})
across layouts. Additional fixed-tilt comparisons,
sensitivity results, and adaptation plots for the other
layouts are provided in the supplementary material. Figure~\ref{fig:paper_scenario_tl}(c) shows the effect
of tightening in the \emph{top-left} layout. Even
without tightening ($\gamma=0$), TCA-DPC achieves
$94\%$ goal-and-constraint success, increasing to
$100\%$ at $\gamma=0.5$\,cm and all larger displayed
values. Without tightening, it also achieves $80\%$
and $74\%$ success in the \emph{top-right} and
\emph{both} layouts, respectively, with higher mean
G\&C than both DPCC variants in each case. These
additional results are reported in the supplementary material.  
Recorded mean computation times per control step are
$1.0$\,s for TCA-DPC and approximately $0.7$\,s for
DPCC. The additional cost arises from repeated denoising,
constraint evaluation, and gradient computations within
MALA. TCA-DPC nevertheless avoids projection and
can operate without constraint tightening ($\gamma=0$).
Reducing this computational cost is a direction for
future work.
\vspace{-1 em}

\section{Conclusion}\label{sec:conclusion}

We proposed a projection-free method that adapts a
pretrained diffusion prior to test-time safety constraints
through exponential tilting, MALA, and
verification feedback. D3IL experiments showed improved
goal-and-constraint success over untilted baselines,
including without constraint tightening.

\clearpage 

\bibliography{l4dc2026-sample}

\newpage

\appendix
\renewcommand{\presectionnum}{}
\startcontents[supplementary]
\section*{Supplementary contents}
\printcontents[supplementary]{}{1}[2]{}
\bigskip

\section{Proof for Proposition~\ref{prop:tilt}} \label{app:sec_proof}
Since $\Phi_t(Y)=0$ on $\mathcal A_t$, for every measurable
$A\subseteq\mathcal Y$,
\begin{equation*}
\begin{split}
 \mathbb Q_{\beta,t}(A\mid\mathcal A_t)
 &=
 \frac{\mathbb Q_{\beta,t}(A\cap\mathcal A_t)}
      {\mathbb Q_{\beta,t}(\mathcal A_t)}
 =
 \frac{\mathbb P_t(A\cap\mathcal A_t)/\mathcal Z_{\beta,t}}
      {\mathbb P_t(\mathcal A_t)/\mathcal Z_{\beta,t}}
 =
 \mathbb P_t(A\mid\mathcal A_t).
 \end{split}
\end{equation*}
This proves preservation of the conditional distribution. Furthermore,
\(
 \mathcal Z_{\beta,t}
 =
 \mathbb P_t(\mathcal A_t)
 +
 \int_{\mathcal A_t^c}
 e^{-\beta\Phi_t(Y)}\,\mathbb P_t(dY).
\)
The integral is non-increasing in $\beta$ and converges to zero
by dominated convergence, since $\Phi_t>0$ on $\mathcal A_t^c$.
Thus, $\mathcal Z_{\beta,t}\downarrow
\mathbb P_t(\mathcal A_t)>0$, which gives
\(
 \mathbb Q_{\beta,t}(\mathcal A_t^c)
 =
 1-\frac{\mathbb P_t(\mathcal A_t)}{\mathcal Z_{\beta,t}}
 \ \downarrow\ 0
 \, \text{as }\beta\to\infty.
\)

\section{Implementation details} \label{sec:implementation_details}

\paragraph{Sequences, constraint residuals, and penalty.}
Each of the $N=8$ stages of $Y$, the sequence generated by the diffusion model, contains the predicted current and desired end-effector positions, $\hat s_{k\mid t}=(p_{k\mid t},p^{\mathrm{des}}_{k\mid t})$, and the input $\hat u_{k\mid t}$. The residuals $g_{\ell,t}$ comprise, for $k=1,\ldots,N-1$ (the initial stage is fixed to $\hat s_{0\mid t}=s_t$ and not checked), \(
a_j^\top p_{k\mid t}-b_j+\gamma,\, (r+\gamma)^2-\|p_{k\mid t}-c\|^2\), for the halfspaces ($\|a_j\|=1$) and the circle, together with the input bounds for $k=0,\ldots,N-1$, given by the smallest box that contains all demonstrated inputs, and the consistency with an explicit Euler model in which both the current and desired positions integrate the commanded velocity, $p_{k+1\mid t}=p_{k\mid t}+\Delta t\,\hat u_{k\mid t}$ and $p^{\mathrm{des}}_{k+1\mid t}=p^{\mathrm{des}}_{k\mid t}+\Delta t\,\hat u_{k\mid t}$ with $\Delta t=1$ control step, for $k=0,\ldots,N-2$, with residual $r^{\mathrm{dyn}}_{k\mid t}$ being the difference of the two sides. All residuals are in physical units (metres). The penalty is $\Phi_t(Y)=\sum_\ell\rho(g_{\ell,t}(Y)+m)$ with margin $m=10^{-3}$ and $\rho(a)=0$ for $a\le0$, $a^2/(2\eta)$ for $0<a\le\eta$, and $a-\eta/2$ for $a>\eta$, where $\eta=10^{-3}$ ($\rho$ is $C^1$, where $C^1$ is set of all continuously differentiable function).

\paragraph{Diffusion prior and adapted sampler.}
The prior is the pretrained diffusion model: a DDPM with a 1D temporal U-Net~\citep{pmlr-v162-janner22a}, $K=20$ denoising steps, and sequences of $N=8$ stages containing states and inputs. We train five models with different seeds on the demonstrations for $10^5$ steps with batch size $32$ and learning rate $10^{-4}$, use the checkpoint with the lowest test loss, and keep the models fixed during deployment. At each control step we generate $B=4$ sequences. One of the $B$ noise vectors is drawn afresh and the others are reused from the previous control step. Each is refined with $M=5$ MALA steps of size $h=0.05$. A sequence passes verification when its largest state-constraint residual is at most $\epsilon=10^{-5}$, its largest dynamics residual satisfies $\|r^{\mathrm{dyn}}_{k\mid t}\|_\infty\le\epsilon_{\mathrm{dyn}}=0.025$\,m, and its largest input-bound residual is at most $\epsilon_u=3\cdot10^{-3}$. Stage $k=0$ is excluded from the state check because $\hat s_{0\mid t}=s_t$ is imposed rather than generated, so it cannot be changed by refinement. The state constraints encode safety and are therefore verified strictly; a looser input tolerance admits the small overshoots of the planned velocities that the prior produces near the goal. The verification tolerances and the margin used in the
penalty are implementation choices applied to the
constraint residuals. They are distinct from the geometric
tightening parameter $\gamma$, which shifts halfspace
boundaries and enlarges circular obstacles. In the
experiments, the failure fraction is computed using
these verification tolerances. Proposition~\ref{prop:tilt}
concerns the exact zero-penalty set of the theoretical
formulation; the implemented acceptance test need not
coincide with this set. Among the verified sequences, we select the minimizer of $J_{\mathrm{sel}}(Y)=\|(\hat s_{k\mid t}-\hat s^{\mathrm{prev}}_{k+1})_{k=0}^{N-2}\|_2$ over normalized states, where $Y_{\mathrm{prev}}$ always has $N$ stages (a reused sequence is shifted, repeating its last stage); at the first step, the first verified sequence is taken. The tilt strength starts at $\beta_0=10$ and is updated by~\eqref{eq:beta_update} with $\kappa=500$, $\delta=0.25$, and $\beta_{\max}=10^4$.

\clearpage

\section{Additional results and ablation study}\label{app:supp_ablation}

\subsection{Additional results}\label{subsec:additional_results}

\begin{table}[H]
\centering
\caption{Closed-loop results at $\gamma=2.5$\,cm, mean ($\pm$ s.e.) over 5 models $\times$ 10 episodes.
Viol., Steps and FB are averaged over layouts.
Proj.: run-time projection. \textbf{Bold}: best; \underline{underlined}: best without projection.}
\label{tab:main25}
\small
\setlength{\tabcolsep}{3.2pt}
\begin{tabular}{@{}lc cc cc cc ccc@{}}
\toprule
 & & \multicolumn{2}{c}{Top-left} & \multicolumn{2}{c}{Top-right} & \multicolumn{2}{c}{Both} & & & \\
\cmidrule(lr){3-4}\cmidrule(lr){5-6}\cmidrule(lr){7-8}
Method & Proj. & G & G\&C & G & G\&C & G & G\&C & Viol.\,$\downarrow$ & Steps\,$\downarrow$ & FB \\
\midrule
Prior                          & \ding{55} & 1.00 & 0.10{\scriptsize$\pm$.03} & 1.00 & 0.06{\scriptsize$\pm$.02} & 1.00 & 0.02{\scriptsize$\pm$.02} & 16.66 & 74.8 & -- \\
Prior + verification           & \ding{55} & 0.98 & 0.50{\scriptsize$\pm$.09} & 0.70 & 0.14{\scriptsize$\pm$.04} & 0.94 & 0.38{\scriptsize$\pm$.11} & 6.75 & 74.4 & 0.31 \\
\textbf{TCA-DPC}                  & \ding{55} & 1.00 & \textbf{1.00}{\scriptsize$\pm$.00} & 0.98 & \underline{0.98}{\scriptsize$\pm$.02} & 0.94 & \underline{0.90}{\scriptsize$\pm$.04} & \underline{0.07} & \underline{68.1} & 0.13 \\
\midrule
DPCC-C                         & \ding{51} & 1.00 & \textbf{1.00}{\scriptsize$\pm$.00} & 1.00 & \textbf{1.00}{\scriptsize$\pm$.00} & 1.00 & \textbf{1.00}{\scriptsize$\pm$.00} & \textbf{0.00} & \textbf{63.8} & -- \\
DPCC-T                         & \ding{51} & 1.00 & \textbf{1.00}{\scriptsize$\pm$.00} & 1.00 & 0.96{\scriptsize$\pm$.02} & 0.94 & 0.94{\scriptsize$\pm$.06} & 0.04 & 64.3 & -- \\
\bottomrule
\end{tabular}
\end{table}

Table~\ref{tab:main25} reports results with constraint
tightening $\gamma=2.5$\,cm. TCA-DPC achieves $100\%$
and $98\%$ goal-and-constraint success in the
\emph{top-left} and \emph{top-right} layouts, respectively,
comparable to the projection-based methods. The gap
is larger in the more restrictive \emph{both} layout,
where TCA-DPC achieves $90\%$, compared with $100\%$
for DPCC-C and $94\%$ for DPCC-T. Across all layouts,
TCA-DPC substantially improves upon the untilted
verification baseline. The relative performance depends on constraint tightening.
Without tightening ($\gamma=0$), TCA-DPC achieves higher
mean goal-and-constraint success than both DPCC variants
in every layout (Table~\ref{tab:main}). Introducing
$\gamma=2.5$\,cm improves both methods, with a larger
gain for DPCC-C. This is consistent with the role of
tightening in compensating for prediction errors:
projection ensures nominal feasibility, but trajectories
close to constraint boundaries remain vulnerable to
differences between predicted and executed states.
TCA-DPC therefore demonstrates effective adaptation
without a buffer, while DPCC-C achieves higher success
when the tested buffer is included.

\begin{figure}[H]
  \centering
  \includegraphics[width=0.65\linewidth]{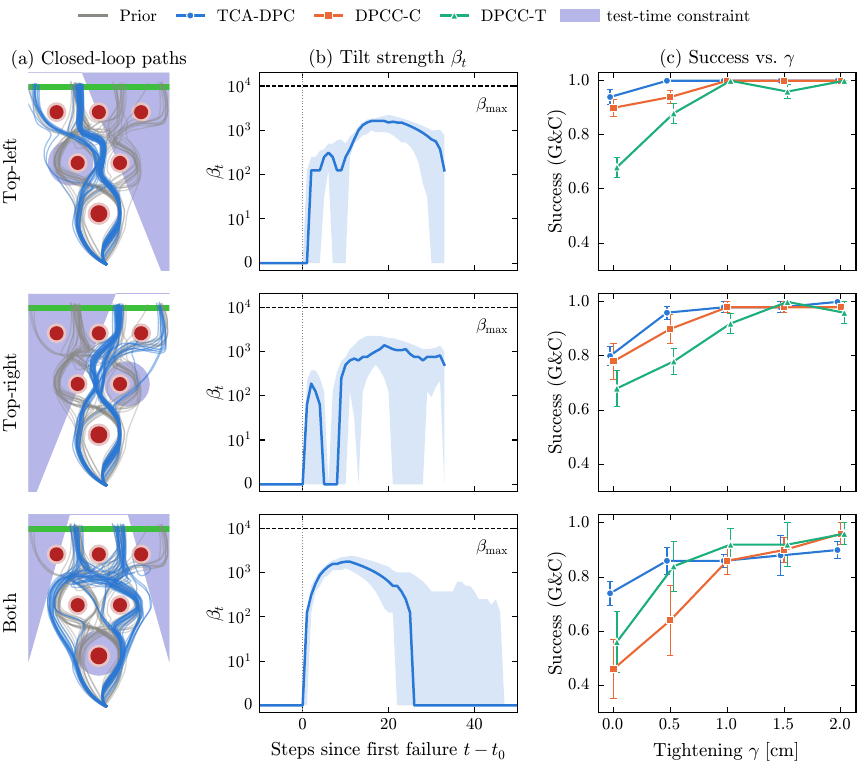}
  \caption{Results on the three constraint layouts, one per row. (a) Closed-loop paths of the prior (grey) and TCA-DPC (blue) without tightening. (b) Tilt strength $\beta_t$ after the first verification failure, median and interquartile range. (c) Success rate versus tightening $\gamma$. Results over 50 episodes per layout.}
  \label{fig:paper_scenarios}
\end{figure}

Figure~\ref{fig:paper_scenarios} extends the comparison
in the main paper to all three test-time constraint
layouts. Column~(a) shows closed-loop trajectories
without tightening ($\gamma=0$): TCA-DPC trajectories largely avoid
both the original obstacles and the additional forbidden
regions, whereas trajectories generated by the prior
frequently enter the latter. Column~(b) shows that, in every
layout, the tilt strength rises to the order of
$10^2$--$10^3$ after the first verification failure, well
below $\beta_{\max}$, and decreases once failures become
rare, most clearly in the \emph{both} layout. Column~(c)
shows that, without tightening ($\gamma=0$), TCA-DPC achieves
higher mean goal-and-constraint success than both DPCC
variants in every layout. The largest difference occurs
in the \emph{both} layout, with $74\%$ success for
TCA-DPC, compared with $46\%$ for DPCC-C and $56\%$
for DPCC-T.

\subsection{Ablation study}
On the \emph{both} layout at $\gamma=2.5$\,cm, each row of
Table~\ref{tab:ablation} changes one parameter from the reference
setting (Section~\ref{sec:implementation_details}), except the
fixed-tilt row, which uses $\beta=10^3$, and the $M=0$ row, which
also sets $M_{\mathrm{extra}}=0$. Most of the gain
comes from the first few MALA steps, with little further
improvement beyond $M=5$. In this experiment, adaptive tilting
attains higher mean success than a fixed $\beta=10^3$
($\kappa=0$), and, for $\kappa\in\{100,500,2000\}$, mean
success increases with $\kappa$.
Success is robust to $\delta\in\{0.05,0.25,0.5\}$, with
$\bar f\approx0.25$ in all cases. Raising $\beta_{\max}$ did
not change the reported performance.
\begin{table}[H]
\centering
\caption{Ablations on the \emph{both} layout at $\gamma=2.5$\,cm. $\bar f$: mean failure fraction.}
\label{tab:ablation}
\small
\setlength{\tabcolsep}{5pt}
\begin{tabular}{@{}llcccc@{}}
\toprule
Parameter & Value & G\&C\,$\uparrow$ & Viol.\,$\downarrow$ & $\bar f$ & FB \\
\midrule
\multicolumn{2}{@{}l}{Reference (TCA-DPC)} & 0.90{\scriptsize$\pm$.04} & 0.10 & 0.25 & 0.17 \\
\midrule
MALA steps $M$ & 0 (prior + verification) & 0.38{\scriptsize$\pm$.11} & 1.84 & 0.28 & 0.24 \\
 (ref.\ 5)     & 1  & 0.78{\scriptsize$\pm$.08} & 0.26 & 0.24 & 0.18 \\
               & 2  & 0.76{\scriptsize$\pm$.05} & 0.50 & 0.26 & 0.19 \\
               & 10 & 0.88{\scriptsize$\pm$.06} & 0.12 & 0.26 & 0.18 \\
\midrule
Gain $\kappa$  & 0 (fixed $\beta=10^3$) & 0.84{\scriptsize$\pm$.08} & 0.32 & 0.26 & 0.19 \\
 (ref.\ 500)   & 100  & 0.76{\scriptsize$\pm$.07} & 0.62 & 0.25 & 0.20 \\
               & 2000 & 0.94{\scriptsize$\pm$.02} & 0.00 & 0.25 & 0.16 \\
\midrule
Budget $\delta$ & 0.05 & 0.96{\scriptsize$\pm$.02} & 0.06 & 0.24 & 0.17 \\
 (ref.\ 0.25)   & 0.5  & 0.94{\scriptsize$\pm$.04} & 0.14 & 0.25 & 0.18 \\
\midrule
Cap $\beta_{\max}$ (ref.\ $10^4$) & $10^5$ & 0.90{\scriptsize$\pm$.04} & 0.10 & 0.25 & 0.17 \\
\bottomrule
\end{tabular}
\end{table}



\end{document}